\documentclass[letterpaper, 10 pt, conference]{ieeeconf}
\usepackage[utf8]{inputenc}
\IEEEoverridecommandlockouts
\DeclareUnicodeCharacter{2212}{-}
\pdfoutput=1 

\usepackage[british]{babel}

\usepackage[T1]{fontenc}
\usepackage{newtxtext}

\usepackage{graphicx}
\graphicspath{{figures/}}

\usepackage{amsmath,amssymb,amsfonts}

\usepackage[hyphens]{url}
\usepackage[colorlinks=true, allcolors=blue]{hyperref}
\usepackage{balance}

\title{\LARGE \bf
C-3TO: Continuous 3D Trajectory Optimization on Neural Euclidean Signed Distance Fields}
\author{
Guillermo Gil$^{1}$, Jose Antonio Cobano$^{1}$, Luis Merino, \textit{Member, IEEE}$^{1}$ and Fernando Caballero$^{1}$%
\thanks{$^{1}$Service Robotics Laboratory -- Universidad Pablo de Olavide (Seville), Spain
{\tt\small \{ggilgar, jacobano, lmercab, fcaballero\}@upo.es}}
}

\begin{document}

\maketitle
\thispagestyle{empty}
\pagestyle{empty}

\begin{abstract}

This paper introduces a novel framework for continuous 3D trajectory optimization (C-3TO) in cluttered environments, leveraging online neural Euclidean Signed Distance Fields (ESDFs). Unlike prior approaches that rely on discretized ESDF grids with interpolation, our method directly optimizes smooth trajectories represented by fifth-order polynomials over a continuous neural ESDF, ensuring precise gradient information throughout the entire trajectory. The framework integrates a two-stage nonlinear optimization pipeline that balances efficiency, safety and smoothness. Experimental results demonstrate that C-3TO produces collision-aware and dynamically feasible trajectories. Moreover, its flexibility in defining local window sizes and optimization parameters enables straightforward adaptation to diverse user’s needs without compromising performance. By combining continuous trajectory parameterization with a continuously updated neural ESDF, C-3TO establishes a robust and generalizable foundation for safe and efficient local replanning in aerial robotics. The source code is open source and can be found at: \url{https://github.com/robotics-upo/icuas2026_neural_trajectory_planner}
\end{abstract}

\section{Introduction} \label{sec:intro}



Aerial robots have become increasingly popular for a wide range of real-world applications due to their ability to perform hazardous tasks more efficiently and, most importantly, more safely than humans \cite{fr21mbzirc}\cite{Ullah2023}. 
Fast trajectory replanning remains a critical area of research, particularly in dynamic and unstructured environments. Equally important is maintaining a continuously updated representation of the drone’s surroundings, which is essential for generating continuous, safe, and smooth 3D local trajectories in real time. This paper presents a framework for planning a continuous local trajectory on an online, neurally-generated, distance field.



Chosing an adequate map representation is key. Having an efficient calculation of the free space and the direction to the closest obstacle are some of those desirable features for such representation. Euclidean Signed Distance Fields (ESDFs) have become increasingly popular as a method for representing and modeling robot surroundings for planning purposes \cite{FIESTA,Oleynikova2016} 
and present many very useful properties that are desirable for planning: it is continuous, differentiable everywhere except at the cut-locus, its gradient is Lipschitz-continuous everywhere except at the cut-locus, and the norm of its spatial gradient is one \cite{Jones:TVCG2006}. Representations of ESDFs are typically based on discrete voxel grids or neural networks. Discrete methods for online ESDF representation, such as Voxblox \cite{voxblox_2017}, FIESTA \cite{FIESTA}, and Voxfield \cite{pan2022}, have gained popularity, but require interpolation to produce continuous ESDF values. In contrast, neural networks can represent ESDFs in a continuous manner 
\cite{Ortiz:etal:iSDF2022}
. HIO-SDF \cite{vasilopoulos_hiosdf_2024} introduced an incremental, online, and global ESDF model represented by a Sinusoidal Representation Neural Network (SIREN) \cite{sitzmann2020implicitneuralrepresentationsperiodic}
. Compared to iSDF \cite{Ortiz:etal:iSDF2022}, HIO-SDF can capture finer details, producing smoother surfaces and incorporating more geometric information throughout the environment. There are other neural approaches such as \cite{maric2024}, where the network learns implicit representations using piecewise polynomial basis functions, or ReDSDF \cite{liu2022}, which uses inductive bias during training to enhance the resulting neural representation. 


Regarding path and trajectory planning for drones, the traditional approaches are sampling-based or searching-based planners
. They can generate optimal paths, but overlook path safety, which makes them undesirable for real-world operations in occluded spaces without post-processing. There are several state-of-the-art entries showing that ESDFs can be very convenient and useful tools for path planning methods \cite{Oleynikova2016,iros2022_edf,FELP,Tordesillas_2019}. Heuristic search planners, by integrating ESDFs and leveraging their properties, can inherently address the safety problem and have demonstrated the capability to compute feasible, safe, and fast paths. However, these discretized paths, defined by a sequence of intermediate waypoints, are neither continuous nor smooth, and do not take into account kinematic or dynamic constraints \cite{gil2025}. 
The next step is rajectory planning through non-linear optimization. It results in trajectories compliant with complex restrictions, assessed in the form of cost functions and easily customized depending on the needs of the user. 
Although trajectory replanning has been widely investigated, most methods depend on discretized ESDFs with interpolation, limiting gradient accuracy and trajectory quality. 

This work focuses on performing local continuous 3D trajectory planning using non-linear optimization directly on an online generated neural ESDF. We present a framework that starts by building the drone's environment representation using 3D LIDAR measurements to train a SIREN-like network, based on the network described in \cite{vasilopoulos_hiosdf_2024}. The framework then leverages the properties of the neural representation to perform trajectory optimization on that online ESDF, taking into account distance to obstacles as an indication of safety in addition to other restrictions.

The main contribution of the proposed framework is to maintain a level of computational optimization sufficient for it to be suitable for use in trajectory replanning, while highlighting its robustness and flexibility compared to existing approaches. The novelty of the framework lies in the optimization of continuous trajectories over continuous ESDFs that are updated online. 

We have organized this paper into six sections. Section \ref{sec:relat} describes the current state of the art
. Section \ref{sec:frame} provides an overview of the framework implemented. Section \ref{sec:path} provides a detailed description of the trajectory planner on the neural ESDF. The experimental validation 
can be found in Section \ref{sec:exper} and conclusions are presented in Section \ref{sec:conc}.

\section{Related Work} \label{sec:relat}

In a taxonomy specific of continuous 3D trajectory planning, the methods could be divided into: hard-constrained methods and gradient-based optimization methods.

Among hard-constrained methods, minimum-snap trajectory approaches are particularly prominent
\cite{Burke_2020}. 
A well-known limitation of these methods lies in the time allocation of the polynomial segments, which can lead to unsatisfactory results. To address this issue, \cite{Ding_ICRA_2018} proposed a real-time B-spline-based kinodynamic search algorithm combined with an elastic optimization procedure as a post-processing step. Another solution is the use of a mixed-integer QP formulation to achieve more suitable time allocations \cite{Tordesillas_2019}.

Gradient-based trajectory optimization methods have proven to be highly effective for efficient local replanning, a critical capability to achieve high-speed flight in unknown environments \cite{Fu_2024,ego-planner}.
In these approaches, the trajectory generation problem is formulated as a non-linear optimization task that balances smoothness, safety, and dynamic feasibility. They can use the ESDF map for obstacle avoidance \cite{Zhou2019} \cite{Tordesillas_2019}. The authors of \cite{Tordesillas_2019} propose a MIQP (Mixed-Integer Quadratic Program) formulation that allows the solver to choose the trajectory interval allocation. The problem is solved using Gurobi\footnote{https://docs.gurobi.com/projects/optimizer/en/current/index.html}.

Many works in the literature incorporate gradient information from an ESDF and utilize the convex hull property of B-spline. Table \ref{table_comparisons} shows the characteristics of several of them. EGO-Planner \cite{ego-planner} avoids maintaining an ESDF entirely, instead extracting obstacle information only for colliding segments and relying on a guide trajectory. Oleynikova et al. \cite{Oleynikova2016} pioneer continuous-time trajectory optimization with high-degree polynomial and Hermetian splines, but their optimization operates on discrete occupancy maps with interpolated ESDFs. Fast-planner method \cite{Zhou2019} decouple the online fast motion planning problem as a front-end kinodynamic path search and a back-end nonlinear trajectory optimization. Improves the smoothness and clearance of the trajectory by a B-spline optimization that incorporates gradient information from an ESDF. The trajectory optimization is solved by a general non-linear optimization solver NLopt\footnote{https://nlopt.readthedocs.io/en/latest/}. Raptor method \cite{raptor} proposes a path-guided optimization approach that incorporates multiple topological paths that represent trajectories as B-splines. It uses \cite{FIESTA} to generate the ESDF. EWOK \cite{Usenko2017} introduces a trajectory replanning framework based on uniform B-splines and a 3D circular buffer, where collision costs are obtained from a discrete Euclidean Distance Transform (EDT) within a local window of about $3m$ and limited to a safety threshold of $0.5m$.


In contrast to traditional approaches that obtain the ESDF via interpolation, an analytical and continuous formulation of the ESDF provides smooth gradients that can be computed efficiently, which greatly facilitates real-time trajectory optimization. On the other hand, compared to B-spline representations, continuous-time polynomial trajectories offer the advantage of including cost functions in order to evaluate the trajectory, rather than only at discrete control points. This continuous formulation allows for a more accurate optimization of smoothness, dynamic feasibility, and safety, making polynomials particularly suitable when high precision in trajectory optimization is required. Moreover, B-splines sacrifice precise control over waypoints and the smoothness of higher-order derivatives.

Therefore, new approaches are required to fully exploit the benefits of continuous ESDFs in trajectory optimization, allowing flexible local map definitions and user-configurable safety margins for planning. Furthermore, a continuous ESDF provides a more consistent and continuous gradient that could avoid oscillations or trajectories with “spikes” that sometimes appear, for example, with heuristic methods.


\begin{table*}[ht]
\centering
\caption{Characteristics of trajectory planning methods.}
\vspace{-2mm}
\begin{tabular}{|p{1.8cm}|p{2.1cm}|p{2.1cm}|p{2.1cm}|p{2.1cm}|p{2.1cm}|p{2.1cm}|}
\hline
\textbf{Characteristics} & \textbf{EWOK \cite{Usenko2017}} & \textbf{EGO \cite{ego-planner}} & \textbf{Fast-planner \cite{Zhou2019}} & \textbf{RAPTOR \cite{raptor}} & \textbf{Oleynikova et al.\cite{Oleynikova2016}} \\
\hline
Distance field & Discrete EDT in local window & 
No ESDF & Corridor-based; no continuous ESDF & Perception-aware corridors; discrete obstacle maps & Discrete maps with interpolated SDF \\
\hline
Local window & Fixed $\approx$3 m buffer & Local occupancy grid & Precomputed convex corridors & Dynamic convex corridors & Occupancy grid with interpolation \\
\hline
Safety margin & Fixed threshold 0.5 m & Via guide path & Corridor boundaries & Corridor boundaries + sensor FOV & Limited by discretization and interpolation \\
\hline
Trajectory representation & B-splines & B-splines & Polynomials  & Polynomials & Continuous-time B-splines \\
\hline
Computation & Real-time but buffer-limited & Very low latency; no ESDF update & Efficient but requires corridor precomputation & More complex; overhead from perception constraints & Online optimization; limited by discrete maps 
\\
\hline
\end{tabular}
\vspace{-2mm}
\label{table_comparisons}
\end{table*}

\section{Framework overview} \label{sec:frame}

\begin{figure*}[t!]
\centerline{\includegraphics[width=.85\textwidth]{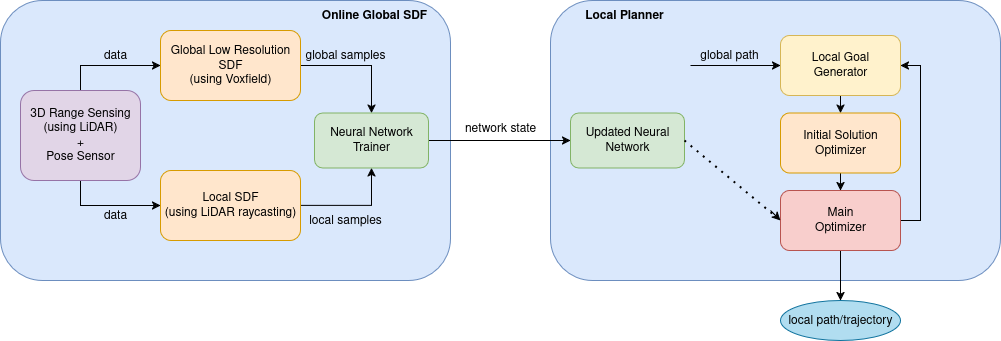}}
\vspace{-5mm}
\caption{Proposed framework structure}
\vspace{-5mm}
\label{architecture_fig}
\end{figure*}

An overview of the entire proposed framework can be found in Fig. \ref{architecture_fig}. It is composed of two different subsystems that work asynchronously to calculate a local trajectory. They are implemented as two independent ROS packages for enhanced modularity.

\subsection{Online Global SDF}

The first subsystem, which we call the Online Global SDF, trains the neural ESDF with the available spatial information. We base its structure on the HIO-SDF solution \cite{vasilopoulos_hiosdf_2024}. First, an onboard 3D LiDAR collects measurements of the drone's surroundings in the form of a local point cloud. This information is then translated into global coordinates using any pose sensor. We assume a perfect localization for the purpose of this paper.

The obtained point cloud is then processed with two complementary methods that obtain coordinate-ESDF pairs, allowing the neural ESDF to be trained. First, a very low-resolution ESDF of the environment that has been already explored is updated and sampled. This allows the system to obtain \emph{global samples} (i. e. samples evenly distributed over the entire known map). This low-resolution ESDF is built online using Voxfield \cite{pan2022}. To obtain \emph{local samples} (i. e. samples corresponding to points near detected obstacles), the second method performs raycasting from the LiDAR location to the points of the obtained point cloud. Local samples are generated by sampling points along the virtual rays and then calculating an approximation of its ESDF considering the LiDAR point cloud. The use of both global and local samples allows the neural network to be trained to predict the best approximation of the ESDF in any given point. A more in-depth description of the sampling methods can be found in \cite{vasilopoulos_hiosdf_2024}.

Subsequently, resulting samples are used to train the desired chosen neural representation of the drone's surroundings. We use a Sinusoidal Representation Neural Network (SIREN) \cite{sitzmann2020implicitneuralrepresentationsperiodic}. It is a fully-connected neural network with 4 hidden layers of 256 neurons each. Its name refers to the use of sinusoidal activation layers. We have chosen this configuration, as it has been shown to offer better results than other approaches, such as the one proposed in \cite{Ortiz:etal:iSDF2022}, specifically for spatial representation \cite{vasilopoulos_hiosdf_2024}. However, the second subsystem of the framework is prepared to work with any other network structure that the user inputs, as explained below.

Tuning the network training parameters is not trivial. In \cite{gil2025} it is shown that to allow the original HIO-SDF network to be suitable for path planning, global samples had to be significantly increased. This reduces network noise in the free-space, which is especially important for gradient calculation.

\subsection{Online Local Planner}

The second subsystem in the framework is responsible for calculating the desired continuous trajectory. As Figure \ref{architecture_fig} shows, the second subsystem expects the first to send a copy of the updated network state, to have the most up-to-date version of the ESDF for it to be used during the planning loop. The first subsystem stores all the neural network information into a .pt file each time a new training loop is performed. This file stores not only the network parameters, but also its entire structure. This means that if the network is changed in future versions of the framework, the second subsystem will not be affected. The first subsystem sends a message every time a new model is available. This allows for asynchronous operation. This is needed since the network training rate is much lower than the planning loop operation rate, for optimization reasons.

The planning loop waits for a global path and the first network version to start its operation. Once both conditions are met, the loop starts by calculating a local goal. The framework builds a 3D local window, whose dimensions can be adapted to the user's needs, and selects the furthest global waypoint that is contained inside that local window. To avoid selecting an invalid or unsafe point as the current local goal, it checks that the desired point has a safe ESDF value, choosing other waypoints otherwise.

Once a local goal is defined, the local trajectory is calculated using two serial optimization processes. Section \ref{sec:path} explores why this is the most desirable approach and provides an in-depth analysis of the optimizers. 

The final result is a local continuous trajectory in the form of three fifth-order polynomials, one for each spatial axis. We chose fifth-order polynomials because they combine the flexibility of high-order polynomials with the low number of parameters to be optimized (which impacts computation time) of lower-order functions.

After the loop starts, it continues to plan new trajectories without waiting for new neural network states, always using the most up-to-date version provided by the training subsystem. In \cite{gil2025} it is shown that planning multiple times with the same network state yields good results and does not pose a greater risk to drone safety when using discrete heuristic path planners over a discretized ESDF. In Section \ref{sec:exper} we show that the statement is also true for the continuous trajectory planning over continuous ESDFs that we propose.

\section{Trajectory Planning on ESDFs} \label{sec:path}

This section details the proposed trajectory planning optimizers. We have chosen a two-step approach, serializing two different optimizers and taking the final state of the first one as the first state of the second one. For the optimization we use Ceres Solver\footnote{http://ceres-solver.org/}, an open source C++ library.

\subsection{Initial Solution Optimizer} \label{subsec:initial}

The first optimizer aims to calculate an initial solution based on the global path waypoints that fall within the local window. The experimental validation in Section \ref{subsec:opt} shows that this initial step helps the main optimizer to converge faster, effectively reducing the total computation time.


The optimizer models the trajectory as the following fifth-order polynomial as a function of the independent variable $\tau$ (time) represented in parametric form:
\begin{align}
x(\tau) &= c_{0}\tau^{5} + c_{1}\tau^{4} + c_{2}\tau^{3} + c_{3}\tau^{2} + c_{4}\tau + d_{0} \\
y(\tau) &= c_{5}\tau^{5} + c_{6}\tau^{4} + c_{7}\tau^{3} + c_{8}\tau^{2} + c_{9}\tau + d_{1} \\
z(\tau) &= c_{10}\tau^{5} + c_{11}\tau^{4} + c_{12}\tau^{3} + c_{13}\tau^{2} + c_{14}\tau + d_{2}
\end{align}

Such polynomials are referred to the local reference system, where the drone remains centered in the previously defined local window. This causes coefficients $d_0$, $d_1$ and $d_2$ to be fixed beforehand. Thus, we define the state vector $\mathbf{c}$ of our optimization problems as the one that contains all non-constant coefficients.
\begin{equation}
\mathbf{c} = 
\begin{bmatrix}
c_0, c_1, c_2, \dots, c_{12}, c_{13}, c_{14}
\end{bmatrix}
\label{eq:state}
\end{equation}

To avoid ill-conditioned optimization problems, we normalize the planning time $\tau \in [0, 1]$ with $\tau = \frac{t}{t_{goal}}$. This prevents large disparities in the scale of polynomial basis functions, which would compromise numerical stability and convergence during the optimization process.

The initial optimizer states are defined for the initial trajectory to be a straight line between the local start and the local goal. This causes all but first- and constant-order coefficients to be zero. Let $(x_{\text{goal}}, y_{\text{goal}}, z_{\text{goal}})$ be the local coordinates of the local goal, the three non-zero initial coefficients can be calculated as follows.
\begin{equation*}
c_4' = \frac{x_{\text{goal}} - d_0}{\tau_{goal}}, \quad
c_9' = \frac{y_{\text{goal}} - d_1}{\tau_{goal}}, \quad
c_{14}' = \frac{z_{\text{goal}} - d_2}{\tau_{goal}}
\end{equation*}

The initial state $\mathbf{c'}$ is then easy to define.
\begin{equation*}
\mathbf{c'} = 
\begin{bmatrix}
0, 0, 0, 0, c_4', 0, 0, 0, 0, c_9', 0, 0, 0, 0, c_{14}'
\end{bmatrix}
\end{equation*}

Once the initial state is calculated, the iterative optimization process starts. The cost function to be optimized has two cost components. The total cost is calculated by squaring the components $f$ so that both positive and negative values are penalized.

The first cost component tries to adapt the trajectory to the global waypoints that land inside the local window. Let $n_{W}$ be the number of global waypoints inside the local window. Let $d_{i,j}$ be the distance between the waypoints $i$ and $j$. An approximation of the reparametrized time value $\tau_i$ at which the trajectory should pass through the waypoint $i$ can be calculated by computing the Euclidean distances between the waypoints and assuming constant speed between them.
\begin{equation*}
\tau_0 = \tau_{start} = 0, \quad
\tau_i = \tau_{i-1} + \frac{d_{i,i-1}}{\sum_{k=1}^{N_{W}} d_{k,k-1}}
\end{equation*}

Let $(x_i, y_i, z_i)$ be the local coordinates of the waypoint $i$, we can define the function $f_{01}$ that penalizes the trajectory not adjusting to the desired waypoints in the desired timestamps.
\begin{equation}
\begin{split}
f_{01}(\mathbf{c}) &= w_{01} \sum_{i=1}^{N_{W}} \Bigl( (x(\tau_i)-x_i)^2 \\
&\quad + (y(\tau_i)-y_i)^2 + (z(\tau_i)-z_i)^2 \Bigr)
\end{split}
\end{equation}

The second cost component function is a smoothness factor. Large coefficients associated with high-order terms could negatively affect the velocity profile, as it is desirable to maintain close to zero acceleration. Therefore, this function aims to penalize states with large coefficients in terms of order greater than one. 
\begin{equation} \label{ini_smooth}
\begin{split}
f_{02}(\mathbf{c}) &= w_{02} \big( |c_0| + |c_1| + |c_2| + |c_3| + |c_5| + |c_6| \\
&\quad + |c_7| + |c_8| + |c_{10}| + |c_{11}| + |c_{12}| + |c_{13}| \big)
\end{split}
\end{equation}

Both $w_{01}$ and $w_{02}$ are weights that will be discussed in Section \ref{sec:exper}.

This optimizer produces an adequate initial solution that will serve as input for the next optimizer, discussed below.

\subsection{Main Optimizer}

This optimizer focuses on the refinement of the provided initial solution according to four different functions. The trajectory and state representation remain the same as in (\ref{eq:state}). 


The first cost component function penalizes the trajectory length. This ensures that the final solution is as short and efficient as possible. This cost is computed by sampling points along the path and calculating the Euclidean distance between them. Let $n_{len}$ be the number of subdivisions and $\tau_j = \frac{j}{n}$ the value of the independent variable $t$ for a certain sampled point $j$, the function can be written as follows.
\begin{equation}
\begin{split}
f_{11}(\mathbf{c}) &= w_{11}\sum_{j=1}^{n_{len}} \Bigl( (x(\tau_j)-x(\tau_{j-1}))^2 \\
&\quad + (y(\tau_j)-y(\tau_{j-1}))^2 + (z(\tau_j)-z(\tau_{j-1}))^2 \Bigr)
\end{split}
\end{equation}

\noindent We used five subdivisions for the experimental validation, which offers a good balance between computational efficiency and trajectory quality.

The second function tries to maintain a safe distance from all obstacles by evaluating the neural ESDF. As with the first function, our approach is to sample points along the path and evaluate the ESDF value of each of them to calculate the associated costs. 

As the coordinates of the points to be evaluated along the trajectory may change after each optimization step (as the state itself changes), querying the network should be done after each of them. We take advantage of the evaluation callbacks offered by Ceres Solver to evaluate the ESDF values of the sampled points, only querying those that are required in the next optimization step. ESDF gradients are also provided by the network. To overcome the computational bottleneck of querying the network \cite{gil2025}, we parallelize it using the tools integrated in Pytorch. As a whole, our querying algorithm reduces significantly the computational effort, maximizing the loop rate, as shown in Section \ref{sec:exper}.  

Let $n_{ESDF}$ be the number of samples considered for this cost function. Let $o_j$ be the ESDF value of the sampled point $j$ and $\tau_j =\frac{j}{n}$ the value of $\tau$ for a certain sampled point $j$. The cost function is formulated as:
\begin{equation} \label{ESDF_cost}
f_{12}(\mathbf{c}) = \frac{w_{12}}{n_{ESDF}-1} \sum_{j=1}^{n_{ESDF}-1} e^{-\alpha(o_j - \sigma)}
\end{equation}

\noindent Notice that both local start and local goal are not considered in the optimization, as they are fixed. The parameters $\alpha$ and $\sigma$, the threshold distance, enable the user to customize the behavior of the optimizer. Consequently, the cost function (\ref{ESDF_cost}) depends on $\sigma$, enabling the system to favor smaller or larger distances to obstacles. Its values are discussed in Section \ref{subsec:opt}. 

The third one is a smoothness function akin to (\ref{ini_smooth}) that penalizes large coefficients associated with high-order terms.
\begin{equation}
\begin{split}
f_{13}(\mathbf{c}) &= w_{13}\big( |c_0| + |c_1| + |c_2| + |c_3| + |c_5| + |c_6| \\
&\quad + |c_7| + |c_8| + |c_{10}| + |c_{11}| + |c_{12}| + |c_{13}| \big)
\end{split}
\end{equation}

The last cost function is in charge keeping the final goal as close to the previously calculated local goal as possible, as it depends on the optimization state variables. Let $(x_{\text{goal}}, y_{\text{goal}}, z_{\text{goal}})$ be the local coordinates of the local goal, the cost function can be written as follows.
\begin{equation}
\begin{split}
f_{14}(\mathbf{c}) &= w_{14}\Bigl( (x(\tau_{\text{goal}})-x_{\text{goal}})^2 \\
&\quad+ (y(\tau_{\text{goal}})-y_{\text{goal}})^2 + (z(\tau_{\text{goal}})-z_{\text{goal}})^2 \Bigr)
\end{split}
\end{equation}

After the optimization process, the output continuous trajectory can be fed into any desired subsystem as polinomyal functions or as a set of waypoints by sampling.

\section{Experimental Validation} \label{sec:exper}

This section aims to evaluate the proposed framework. First, we present how the weights and parameters were selected. Then we benchmark our proposed framework against other similar integrations.

The scenario used is a model of one of the buildings of the Eindhoven University of Technology \cite{Pauwels2023}. 
The hardware and resources used for all the tests are: AMD Ryzen 7 7735hs CPU 3.20GHz, 32 GB RAM, with dockerized Ubuntu OS 20.04 LTS. The algorithms were developed in C++ and Python, embedded in a ROS Noetic Distribution.


As the size of the local window does not influence on the computation time because the number of queries depends only on the parameters discussed in Section \ref{subsec:opt}, we used a local window size of $6m \times 6m \times 3.2m$ 
for comparison with other works. 

\subsection{Optimizer parameter configuration} \label{subsec:opt}

Every component function $f$ described in Section \ref{sec:path} 
has an associated weight constant term that affects the optimization process. The final tuned weights that are used for the experimentation process are the following:
\begin{equation*}
w_{01} = 10.0, \quad
w_{02} = 1.0
\end{equation*}
\begin{equation*}
w_{11} = 1.0, \quad
w_{12} = 3.0, \quad
w_{13} = 0.1, \quad
w_{14} = 10000
\end{equation*}

When selecting the appropriate values, it is key to consider that 
there is no unique correct solution. Cost selection and balancing should be based on the user's needs and desired cost functions. High customization is one of the main features of our framework.

Starting with the initial optimizer, smoothness in this first step is secondary to the fact that the trajectory should adapt as much as possible to the local section of the global path. Because of that, $w_{01}$ should be bigger than $w_{02}$.

For the main optimizer, further reasoning can be applied to select the weights. Reaching the local goal is what is usually named a "hard restriction", which the optimizer must respect above all other costs. Although the Ceres Solver does not have a native feature to implement this, it can be achieved by selecting a weight orders of magnitude higher than any other. This results in $w_{14} >> w_{11},w_{12},w_{13}$. The other three costs represent the weight of the travel distance, safety, and smoothness in the final trajectory, and must be selected based on the desired specifications.

The numeric values defined above were selected through many test iterations until we reached our desired performance. For the hard restriction to be implemented, we found that a difference of three orders of magnitude is enough for the local goal to be completely fixed. Larger weights could affect the stability of the optimizer without offering better performance.

The cost function defined by (\ref{ESDF_cost}) has two parameters that should be tuned. We fixed $\alpha = 4$ because it showed the best overall results. The parameter $\sigma$ has greater importance, as it defines a threshold distance from which the cost function increases rapidly
. In Section \ref{subsec:performance} we experiment with different $\sigma$ values.

\begin{table*}[t!]
\caption{Performance comparison based on optimizer maximum iterations and number of neural ESDF samples ($\sigma = 1.5$)}
\vspace{-3mm}
\begin{center}
\begin{tabular}{|c|c|c|c|c|c|c|c|c|}
\hline
\multicolumn{3}{|c|}{\textbf{Optimizer parameters}} 
& \multicolumn{3}{c|}{\textbf{Loop time (ms)}} 
& \multicolumn{2}{c|}{\textbf{Dist. to obstacles (m)}}
& \multicolumn{1}{c|}{\textbf{Path length (m)}} \\
\cline{1-9}
\textbf{$iter_{ini}$} & \textbf{$iter_{main}$} & \textbf{$n_{ESDF}$} 
& \textbf{Mean} & \textbf{Min.} & \textbf{Max.} 
& \textbf{Mean} & \textbf{Min.} 
& \textbf{Total} \\
\hline
- & - & 20 (Precalculated ESDF) & 1544  & 1025 & 2071 & 1.00 & 0.72 & 4.39 \\
50 & 50 & 15 & 845 & 819 & 875 & 0.97 & 0.67 & 4.58 \\
\textbf{50} & \textbf{50} & \textbf{5} & \textbf{418} & \textbf{327} & \textbf{576} & \textbf{1.04} & \textbf{0.82} & \textbf{4.54} \\
50 & 30 & 15 & 712 & 561 & 1198 & 0.93 & 0.64 & 4.30 \\
\textbf{50} & \textbf{30} & \textbf{5} & \textbf{341} & \textbf{237} & \textbf{551} & \textbf{0.98} & \textbf{0.73} & \textbf{4.49} \\
50 & 10 & 15 & 278 & 178 & 386 & 0.75 & 0.43 & 4.04 \\
50 & 10 & 5 & 124 & 51 & 184 & 0.71 & 0.46 & 4.08 \\
30 & 50 & 15 & 875 & 712 & 1201 & 0.97 & 0.69 & 4.43 \\
30 & 50 & 5 & 487 & 341 & 763 & 1.04 & 0.85 & 4.74 \\
30 & 30 & 15 & 697 & 498 & 946 & 0.95 & 0.64 & 4.37 \\
\textbf{30} & \textbf{30} & \textbf{5} & \textbf{303} & \textbf{203} & \textbf{423} & \textbf{0.90} & \textbf{0.64} & \textbf{4.42} \\
30 & 10 & 15 & 271 & 129 & 403 & 0.74 & 0.37 & 4.13 \\
30 & 10 & 5 & 110 & 53 & 161 & 0.69 & 0.43 & 3.97 \\

\hline
\end{tabular}
\label{tab_esdf_metrics_perf}
\vspace{-5mm}
\end{center}
\end{table*}

There are other optimization parameters that impact not only the performance and the resulting trajectory calculated by the framework, but also the time consumed by the trajectory planner loop. These parameters are the maximum number of optimization iterations for both optimizers, 
$iter_{ini}$ and $iter_{main}$, and the number of points $n_{ESDF}$ considered in the ESDF cost function \ref{ESDF_cost}. 

For this framework to be used as a local trajectory planner, it is mandatory to achieve the highest possible update rate. We compared performance indicators and loop time metrics while varying those three optimizer parameters to select the ones that achieve better balance between rate and performance. With the same local start and local goal and having an unmapped obstacle between them, we varied the number of iterations of both optimizers and the number of ESDF samples. 
Table \ref{tab_esdf_metrics_perf} shows the main results. The metrics shown were obtained by averaging measurements from 20 optimization loops with each configuration. The first row shows the metrics using a precalculated perfect ESDF without limiting the optimizer iterations, as a way to fix an ideal performance for this experiment.

Many conclusions can be drawn from this table. One of the most important, but subtle, ones is that a lower number of iterations of the initial optimizer ($iter_{ini}$) does not translate to a lower loop time. The main optimizer, as it evaluates the ESDF, consumes a greater amount of time per iteration. If the initial solution is good enough, the main optimizer converges faster, resulting in a reduced time per planning loop.

It should also be noted that a larger number of ESDF samples ($n_{ESDF}$) implies a longer computation time, but not necessarily better results. Depending on the level of occlusion of the environment, it is possible to reduce that parameter without compromising safety. However, the maximum number of iterations of the main optimizer ($iter_{main}$) heavily impacts the quality and safety of the path and should not be reduced below the 30 iteration threshold.

The combinations that offer better results are those marked in bold in Table \ref{tab_esdf_metrics_perf}. They will be used to benchmark the framework against other state-of-the-art similar planners.

\subsection{Performance comparison} \label{subsec:performance} 

\begin{figure*}[t!]
\centerline{\includegraphics[width=.95\textwidth]{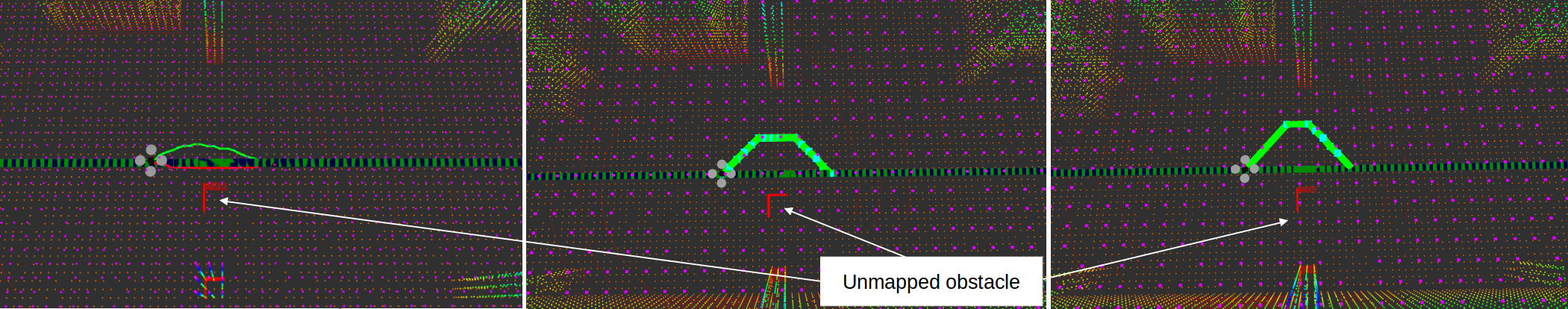}}
\vspace{-2mm}
\caption{$Scenario1$ - local trajectory generated by C-3TO (left), local path generated by the cost-aware A* path planner (center), local path generated by the cost-aware Lazy Theta* path planner (right). An unmapped obstacle is avoided.}
\label{fig:s1_paths}
\end{figure*}

\begin{figure*}[t!]
\centerline{\includegraphics[width=.95\textwidth]{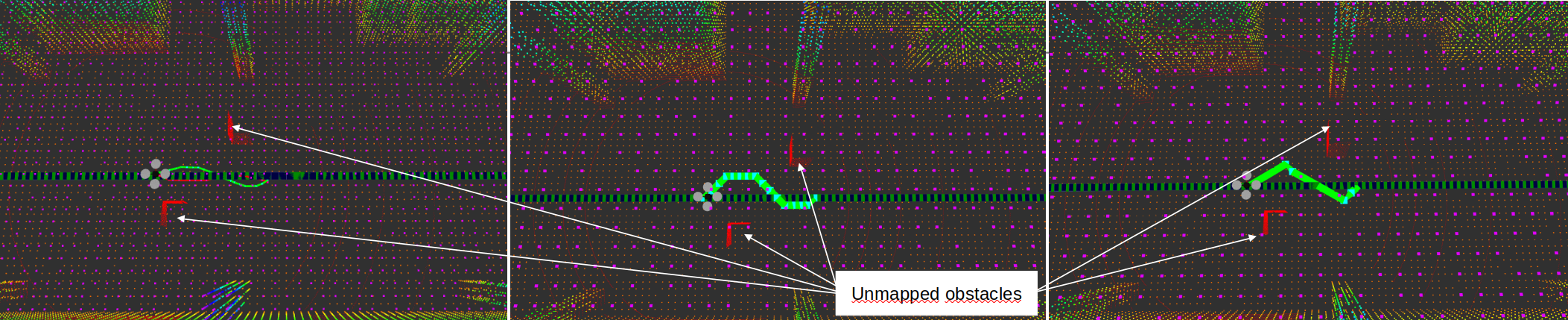}}
\vspace{-2mm}
\caption{$Scenario2$ - local trajectory generated by C-3TO (left), local path generated by the cost-aware A* path planner (center), local path generated by the cost-aware Lazy Theta* path planner (right). Unmapped obstacles are avoided.}
\vspace{-5mm}
\label{fig:s2_paths}
\end{figure*}

This section provides a comparison with other methods in order to demonstrate the capabilities of the framework. The system proposed in \cite{gil2025} is a path planning framework that uses the information provided by a neural network to build a local ESDF grid and perform path planning using heuristic approaches such as A* and Lazy-Theta* over it. We reproduced the same experimental validation performed in \cite{gil2025}, where the drone should face scenarios with unmapped obstacles that should be avoided. Table \ref{icuas_sce1s2} summarizes the performance results for both scenarios described in \cite{gil2025}. Figures \ref{fig:s1_paths} and \ref{fig:s2_paths} illustrate the experimental setup, as well as the results. Table \ref{icuas_sce1s2} compares the three selected parameter configurations of the C-3TO framework with the two algorithms introduced in \cite{gil2025}.

\begin{table*}[t!]
\caption{Performance comparison of different planner and parameter configurations in Scenarios 1 and 2 ($\sigma = 1.5$)}
\vspace{-5mm}
\begin{center}
\begin{tabular}{|c|c|c|c|c|c|c|c|c|c|}
\hline
\textbf{Scenario} & \multicolumn{3}{c|}{\textbf{Parameters / Planner}} 
& \multicolumn{3}{c|}{\textbf{Loop time (ms)}} 
& \multicolumn{2}{c|}{\textbf{Dist. to obstacles (m)}} 
& \multicolumn{1}{c|}{\textbf{Path length (m)}} \\
\cline{2-10}
\textbf{} & \textbf{$iter_{ini}$} & \textbf{$iter_{main}$} & \textbf{$n_{ESDF}$} 
& \textbf{Mean} & \textbf{Min.} & \textbf{Max.} 
& \textbf{Mean} & \textbf{Min.} 
& \textbf{Total} \\
\hline
S1 & 50 & 50 & 5 & 358 & 285 & 531 & 1.16 & 1.11 & 3.16 \\
S1 & 50 & 30 & 5 & 268 & 184 & 374 & 1.04 & 0.90 & 3.22 \\
S1 & 30 & 30 & 5 & 254 & 154 & 362 & 0.94 & 0.88 & 3.14 \\
S1 & \multicolumn{3}{c|}{Cost Aware A* \cite{gil2025}} & 196 & 164 & 216 & 1.36 & 1.14 & 3.97 \\
S1 & \multicolumn{3}{c|}{Cost Aware LazyTheta* \cite{gil2025}} & 204 & 171 & 246 & 1.37 & 1.22 & 3.60 \\
\hline
S2 & 50 & 50 & 5 & 364 & 260 & 516 & 1.10 & 0.82 & 3.28 \\
S2 & 50 & 30 & 5 & 274 & 186 & 400 & 1.06 & 0.85 & 3.28 \\
S2 & 30 & 30 & 5 & 266 & 154 & 402 & 0.97 & 0.88 & 3.15 \\
S2 & \multicolumn{3}{c|}{Cost Aware A* \cite{gil2025}} & 207 & 160 & 308 & 1.10 & 0.70 & 3.64 \\
S2 & \multicolumn{3}{c|}{Cost Aware LazyTheta* \cite{gil2025}} & 197 & 185 & 223 & 1.14 & 0.66 & 3.62 \\
\hline
\end{tabular}
\label{icuas_sce1s2}
\vspace{-2mm}
\end{center}
\end{table*}

\begin{figure*}[t!]
\centerline{\includegraphics[width=.95\textwidth]{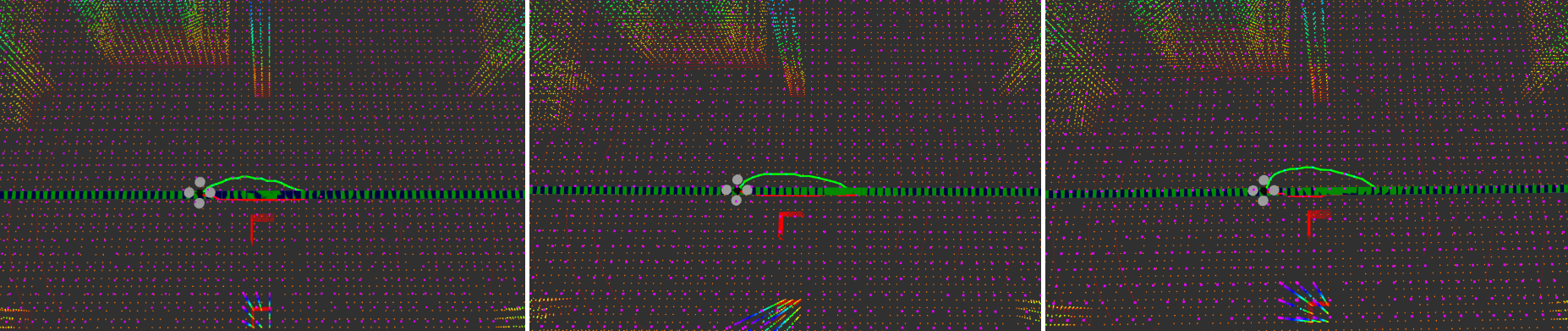}}
\vspace{-2mm}
\caption{$Scenario1$ - local trajectory generated by C-3TO considering $\sigma = 1.5$ (left), $\sigma = 2.0$ (center), $\sigma = 2.5$ (right).}
\vspace{-2mm}
\label{fig:s1_exp}
\end{figure*}

\begin{figure*}[t!]
\centerline{\includegraphics[width=.95\textwidth]{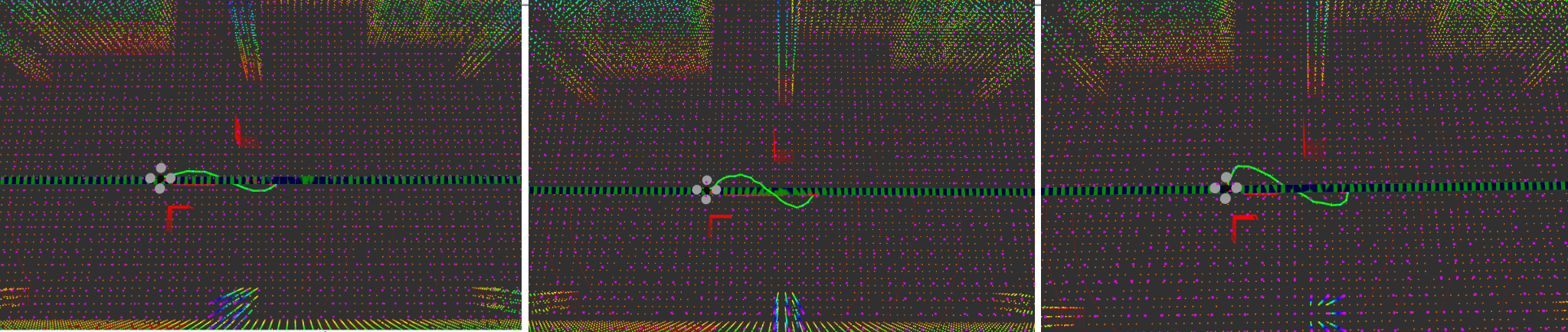}}
\vspace{-2mm}
\caption{$Scenario2$ - local trajectory generated by C-3TO considering $\sigma = 1.5$ (left), $\sigma = 2.0$ (center), $\sigma = 2.5$ (right).}
\vspace{-5mm}
\label{fig:s2_exp}
\end{figure*}

One of the main features is safety. As explained in Section \ref{sec:path}, the ESDF cost function (\ref{ESDF_cost}) depends on the threshold distance $\sigma$. Figures \ref{fig:s1_exp} and \ref{fig:s2_exp} show how the generated trajectory depends on this parameter, increasing the distance to the considered unmapped obstacles with its value.


Smoothness is one of the major differences between the two systems. The framework presented in \cite{gil2025} does not take into account dynamic constraints. However, our system integrates smoothness into the optimization process, resulting in smooth and dynamically aware trajectories (see Fig. \ref{fig:s1_paths} and \ref{fig:s2_paths}). Further kinodynamic constraints can also be easily added to the Ceres optimization environment, should the user need them.

Another improvement of our framework relies on computation time. The system in \cite{gil2025} relies on a volumetric sampling of the ESDF, which causes the loop time to grow cubically with the dimensions of the local window. In contrast, in our framework, increasing the local window would only produce a linear increase in the number of samples in (\ref{ESDF_cost}), but the increase of the computation time is not relevant.

One last topic to take into account is the planner convergence time. The heuristic planners used in \cite{gil2025} do not experience substantial improvement when the resulting local path follows the global path, a common occurrence when there are no unmapped obstacles nearby. However, the two-step optimization of our system allows the main optimizer to converge faster in those cases, as the initial optimizer already outputs the final local path. This greatly decreases the total loop time, as shown in the simulations, where the time metrics decreased to less than half of those shown in Table \ref{tab_esdf_metrics_perf} in those situations.

\subsection{Discussion}

Next, we discuss the advantages of the proposed framework in comparison with the methods shown in Table \ref{table_comparisons}. 
Our framework directly optimizes trajectories on a continuously updated neural ESDF, enabling smooth, dynamically feasible fifth-order polynomials with configurable safety margins. This contrasts with prior methods such as EWOK \cite{Usenko2017}, which relies on discrete EDT grids with fixed windows, EGO-Planner \cite{ego-planner}, which avoids ESDF maintenance by using guide trajectories, and Fast-Planner \cite{Zhou2019} or RAPTOR \cite{raptor}, which depend on convex corridor constructions. Unlike the continuous-time B-spline optimization of Oleynikova et al. \cite{Oleynikova2016}, our approach avoids interpolated distance fields, providing consistent gradients and overcoming the main limitations of discrete, corridor-based, or heuristic planners.



In summary, C-3TO achieves high smoothness through continuous polynomial trajectories, high robustness from globally consistent ESDF gradients, and superior flexibility by allowing user-defined safety margins and local window sizes. Compared to B-spline, corridor-based, or buffer-based methods, it consistently delivers smoother, safer, and more adaptable solutions for real-time replanning.

\section{Conclusion} \label{sec:conc}


We present a novel framework for optimizing 3D continuous trajectories, modeled with three fifth-order polynomials, on a continuous online ESDF neural representation. The results demonstrate substantial improvements over prior approaches in both performance and computational efficiency by directly optimizing continuous trajectories over continuous ESDFs, thereby eliminating the interpolation step required in other methods. Optimizing continuous-time polynomials directly over an online neural ESDF yields precise gradients, better dynamic control, and smoother, safer paths. Moreover, the framework’s flexibility allows straightforward adaptation to diverse user requirements, making it suitable for a wide range of tasks. A key advantage is that adjusting the local window size and optimization parameters does not significantly increase computation time, while still yielding robust, safe, smooth, and efficient solutions. Finally, the use of continuous polynomials offers clear benefits over B-splines, enabling more precise optimization.

As future work, we will focus on implementing more complex kinodynamic constraints and exploring new ways of further optimizing the framework.

\section*{Acknowledgment}

This work was supported by the grants COBUILD (PID2024-161069OB-C31, funded by the Ministry of Science, Innovation and Universities, the Spanish Research Agency, and the European Regional Development Fund, MICIU /AEI /10.13039/501100011033 / FEDER, UE), and PICRAH 4.0 (PLEC2023-010353, funded by the Spanish Research Agency and the Ministry of Science, Innovation and Universities, MCIN /AEI /10.13039/501100011033)

\balance

\bibliographystyle{IEEEtran}
\bibliography{icra2026}

\end{document}